\documentclass[11pt]{article}

\usepackage[final]{acl}
\usepackage{booktabs}
\usepackage{numerica}
\usepackage{amsmath}
\usepackage{multirow}
\usepackage{nameref}
\usepackage{multirow}
\usepackage{makecell}
\usepackage{newfloat}
\usepackage{listings}
\usepackage{times}
\usepackage{latexsym}

\newcounter{promptbox}

\usepackage[T1]{fontenc}
\usepackage[utf8]{inputenc}
\usepackage{microtype}
\usepackage{inconsolata}
\usepackage{graphicx}
\usepackage{tikz}
\usepackage{framed}
\usepackage{mdframed}
\usetikzlibrary{positioning,fit,backgrounds}
\usepackage{xcolor}

\title{You Frame It: How Conceptual Representations Shape LLM Detection and Reasoning about Antisemitism}

\author{Katharina Soemer \\
  Goethe University Frankfurt \\
  \texttt{soemer@soz.uni-frankfurt.de} \\\And
  Helena Mihaljević \\
  Hochschule für Technik und Wirtschaft Berlin \\
  \texttt{mihalje@htw-berlin.de} \\}

\begin{document}

\maketitle

\begin{abstract}
    LLMs enable the integration of external conceptual resources at inference time, creating new opportunities for detecting ideologically and historically complex phenomena such as antisemitism. We investigate how different forms of conceptual grounding affect antisemitism detection and explanation behavior across four state-of-the-art LLMs.    
    Using two expert-annotated datasets, we compare definitional, fine-grained taxonomic, example-augmented, and large-context representations of antisemitism.
    We find that fine-grained taxonomic representations substantially improve recall, while simultaneously reducing precision. Surprisingly, supplying substantially larger conceptual resources yields no additional quantitative benefit. Post-Holocaust antisemitism poses the most persistent challenge across models and configurations.  Analysis of explanations further reveals systematic limitations including overproduction of conceptual references, reliance on lexical cues, overconfidence, and difficulties with subtle or justificatory forms of antisemitism.
    Our findings highlight both the potential and the remaining limitations of conceptually grounded LLMs for antisemitism detection and reasoning.\footnote{Code is  available at: \url{https://github.com/katsoemer/antisemitism-llm-conceptual-representations}.}

\end{abstract}

\noindent\textcolor{red}{Warning: This paper discusses antisemitism and contains uncensored slurs. }
\section{Introduction}
Despite advances in hate speech detection using Large Language Models (LLMs) \cite{gilardi_etal_2023, li_etal_2024}, recent studies indicate that antisemitic content detection remains particularly challenging \cite{steffen_etal_2024, patel-etal-2025}.
Models struggle particularly with implicit, coded, or trope-based rhetoric \cite{mihaljevic_steffen_2022, mendelsohn_etal_2023}, and may reproduce or amplify anti-Jewish biases \cite{adl_report_2025}. 

Antisemitism poses a particular challenge for automated detection due to its conceptual and rhetorical complexity.\footnote{A detailed introduction to the concept of antisemitism and its history is provided in Appendix~\ref{app:antisemitism}.} It operates through a heterogeneous, historically evolving repertoire of stereotypes, conspiracy theories, and narratives, not all of which are explicitly hateful, e.g., claims of Jewish financial power and control \cite{decoding_lexicon_2024}. Even expert annotators show  substantial disagreement \cite{steffen_etal_icwsm_2023}, indicating that LLMs require specific contextual knowledge to capture these ideological patterns. 

The knowledge encoded in LLMs is opaque, fixed at training time, and potentially shaped by the same biases the models are expected to detect. Prompting offers the possibility to inject external conceptual representations at inference time, for example through definitions, taxonomies, or examples. 
While external knowledge infusion has been explored in adjacent areas \cite{yang_comprehensive_2025,melis-etal-2025-modular}, little is known about how different forms of conceptual grounding affect detection and classification of different forms of antisemitism.

In this work, we systematically evaluate the effect of external conceptual representations on antisemitism detection and classification across four state-of-the-art LLMs: Gemini-2.5-Flash, Claude Sonnet 4.6, GPT-5.4, and LLaMA-3.3-70b-instruct. We compare four prompting configurations: a baseline without external conceptual grounding; the \textit{IHRA Working Definition of Antisemitism} (hereafter: \textit{IHRA}) \cite{ihra_2016}; and a fine-grained taxonomy derived from the \textit{Decoding Antisemitism Lexicon} (hereafter: \textit{Lexicon}) \cite{decoding_lexicon_2024}  with and without examples. In addition, we evaluate Gemini using the complete 550-page \textit{Lexicon}. 

Our study makes four main contributions:
(1) We provide the first systematic multi-model evaluation of external conceptual grounding for antisemitism detection and classification. 
(2) We show that compact but fine-grained conceptual representations substantially improve recall, while long contextual resources yield no additional benefit.
(3) We demonstrate that different forms of antisemitism vary strongly in detectability across models and prompting configurations, with Israel-related and post-Holocaust antisemitism posing largest challenges.
(4) We analyze explanation behavior as an indirect reasoning signal, revealing systematic overproduction of conceptual references, shortcut reliance on lexical cues, and mismatches between internal uncertainty and final predictions.

Our findings contribute both to the underexplored area of computational antisemitism research and to the broader design of more reliable LLM-based moderation and monitoring systems. Antisemitism constitutes a particularly demanding test case due to its ideological complexity, historical depth, and frequent implicitness, making it well suited for studying the limits of current LLM reasoning and external knowledge integration.

\section{Related Work}
\label{sec:related_work}
\subsection{LLMs for Detecting Antisemitism}
\label{subsec:related_work_llms}

LLMs achieve strong performance on a range of harmful-content detection tasks, often matching or exceeding traditional classifiers and crowd-based annotation approaches \citep{gilardi_etal_2023,li_etal_2024,roy-etal-2023}. At the same time, prior work consistently reports limitations for implicit, coded, and context-dependent forms of hate speech \citep{kumar_2024,huang_is_2023,hartmann_etal_moderation_2025,elsherief-etal-2021-latent} and antisemitism \citep{mendelsohn_etal_2023}, while exhibiting oversensitivity toward identity-related terms,  controversial topics and ignoring intent \citep{zhang_dont_2024,hartmann_etal_moderation_2025,roy-etal-2023}. 
Benchmarking studies additionally indicate large performance discrepancies for antisemitic content compared to other target groups \citep{balachandran_2024_eureka}.

Prompt-based antisemitism detection currently achieves substantially lower performance than typically reported for hate speech, offensive language or conspiracy theory detection \cite{guo_investigation_2024,roy-etal-2023}. On Twitter data sampled using thematically explicit keywords, prompting open LLMs yields F$_1$ scores up to 0.69 \citep{patel-etal-2025}. On more discourse-diverse datasets, performance is substantially lower, with reported F$_1$ scores around 0.55 even for GPT-4 \citep{pustet_mihaljevic_report6_2024,steffen_etal_2024}. 
These findings suggest that antisemitism remains a particularly difficult target for LLM-based moderation, especially in settings without strong lexical cues.

\subsection{Conceptual Grounding and Prompting}
\label{subsec:related_work_kb}

Recent work has explored integrating external conceptual resources at inference time \citep{yang_comprehensive_2025}. Providing definitional, policy-based, or conceptual context can improve detection of harmful content \citep{atreja_etal_2025,roy-etal-2023} and antisemitism \citep{becker_etal_report6_2024,patel-etal-2025}. Even the conceptualization and operationalization of hate speech substantially affect prompting behavior \citep{melis-etal-2025-modular}, with varying effects across models. In the context of antisemitism, \textit{IHRA} constitutes the dominant definitional framework in both policy and computational research \citep{patel-etal-2025}. However, there is still little research on how alternative or substantially more fine-grained conceptual representations affect LLM detection and reasoning behavior.

Findings regarding the inclusion of examples in prompts remain mixed. Some studies report improvements through few-shot prompting \citep{ghorbanpour-etal-2025-prompting}, while others show strong sensitivity to prompt wording, model family, and example selection \citep{chae_davidson_2026,reynolds_mcdonell_2021}. Existing work therefore suggests that prompting performance depends not only on the amount of contextual information provided, but also on how concepts are structured and represented.

Recent long-context LLMs further enable the inclusion of substantially larger conceptual resources such as books or policy documents at inference time. However, it remains unclear whether increasing the amount of contextual material beyond compact conceptual representations further improves classification and reasoning performance.

\subsection{LLM Reasoning Analysis}\label{subsec:related_work_expl} 

LLM-generated explanations are increasingly explored as a mechanism for improving transparency and human oversight in moderation systems \citep{ma_kou_2023,kolla_etal_2024}. Prior work  suggests that prompting models to explain their decisions can improve classification performance itself \citep{agarwal_mama-memeia_2025,atreja_etal_2025}.
Findings on the practical usefulness of LLM explanations in moderation contexts remain mixed. \citet{huang_is_2023} report that ChatGPT-generated explanations for implicit hate speech are rated as clearer than human-written ones, but not necessarily more informative, while \citet{di_bonaventura_is_2024} find only weak alignment between LLM explanations and human expectations. Related work has further explored grounding explanations in explicit policy references. \citet{kumar_2024}, for example, prompt LLMs to identify relevant community guidelines alongside classification decisions. Their manual error analysis further suggests that a substantial share of apparent false positives reflects annotation ambiguity, whereas false negatives more often stem from genuine reasoning failures.
In this context, it should be noted that evaluating explanation quality and faithfulness remains challenging \citep{di_bonaventura_is_2024,zhong_multimodal_2024}, with  studies  frequently relying on qualitative examinations of explanations \citep{mendelsohn_etal_2023,goyal-etal-2025-momoe}.

Our work builds on this literature in several ways. First, we systematically compare different forms of external conceptual grounding, ranging from compact definitions to fine-grained taxonomies and large-context resources. Second, we focus on antisemitism as a particularly challenging and historically grounded form of harmful content. Third, we analyze explanations as indirect signals of conceptual reasoning by mapping generated references onto antisemitic concepts.

\section{Data and Methods}

\subsection{Datasets}
We use two expert-annotated datasets on online antisemitism that differ  in sampling strategy and annotation granularity, and have served as primary sources of labeled examples in prior computational work \citep{becker_etal_report6_2024,patel-etal-2025}.

The \textbf{Bloomington} dataset \citep{jikeli_etal_2021} consists of English-language Twitter (now X) posts collected via the keywords \textit{jews}, \textit{israel}, \textit{kikes}, and \textit{zionazi*}. 
Annotations follow \textit{IHRA}, consisting of a general definitional paragraph and eleven illustrative categories. Posts labeled as antisemitic are further assigned up to two most salient \textit{IHRA} sections \citep{jikeli2024}. 

The \textbf{Decoding} dataset \citep{decoding_lexicon_2024} contains English-language comments from mainstream social media platforms and news outlets, sampled around public events expected to elicit antisemitic discourse. 
The annotation is based on a fine-grained annotation scheme derived from, but extending,  \textit{IHRA}, comprising 41 codes at the same level of granularity that capture specific stereotypes, narratives, and rhetorical patterns of contemporary antisemitism. The  number of codes assigned to a post labeled as antisemitic is not restricted. 

The annotation scheme also underlies the  \textit{Lexicon} \citep{decoding_lexicon_2024}, a 550-page reference work.\footnote{The original Decoding annotation scheme contains additional, partly hierarchical annotation codes. For the Lexicon, these were consolidated by its authors into a flat taxonomy of 41 concepts, which we adopt throughout this paper.}
Each chapter corresponds to a specific antisemitic concept and provides (i) a conceptual description, (ii) typical linguistic and narrative realizations, and (iii) contemporary examples. We use these chapters as a structured taxonomy of antisemitic concepts.
Table~\ref{tab:dataset_comparison} summarizes the main characteristics of the two datasets.

\begin{table}[t]
\centering
\small
\caption{Comparison of the datasets used in this study.}
\begin{tabular}{p{2.0cm} p{2.2cm} p{2.3cm}}
\toprule
\textbf{Characteristic} & \textbf{Bloomington} & \textbf{Decoding} \\
\midrule

Sources 
& Twitter
& Social media \newline\& news outlets\\

Time span 
& 01/19--04/23 
& 12/20--11/23 \\

Sampling  
& Keyword-based
& Discourse-driven\\

Size 
& 11{,}311 
& 42{,}799 \\

\% Antisemitic
& 1{,}953 (17.3\%) 
& 3{,}375 (7.9\%)  \\

Sub-labels 
& 12 \textit{IHRA} sections 
& 41 \textit{Lexicon} chapters  \\

Sub-labels/post 
& max. 2; avg: 1.3 
& No upper bound; avg. 1.5 \\

Sublabel skew 
& Top 5: $\sim$90\%
& Top 5: $\sim$53\% \\

\bottomrule
\end{tabular}
\label{tab:dataset_comparison}
\end{table}

We map both annotation frameworks (\textit{IHRA} sections and \textit{Lexicon} chapters) to a set of higher-level \textbf{content groups} that capture recurring dimensions of antisemitic discourse. This abstraction enables comparative and interpretable analyses across datasets and prompting configurations. Detailed descriptions, label distributions, and mappings are provided in Appendix~\ref{app:dataset_details}.

Both datasets were preprocessed using a unified pipeline; details are provided in Appendix~\ref{app:preprocessing}.

\subsection{Model Choice}
We evaluate Claude 4.6 Sonnet \citep{anthropic2026claude46}, Gemini-2.5-Flash \citep{gemini_2_5_report}, GPT-5.4 \citep{openai2026gpt54}, and LLaMA-3.3.70b-instruct \citep{llama3_report}, covering both proprietary and open-weight model families commonly examined in recent work on antisemitism detection \citep{becker_etal_report6_2024,adl_report_2025,patel-etal-2025}. 
Gemini-2.5-Flash additionally supports substantially larger input lengths and document-style inputs (e.\,g., PDF files). This enables evaluation of a full-lexicon condition that cannot be replicated across all models due to context limitations.

\subsection{Prompting Configurations and Evaluation}

We use instruction-based prompting. All prompt configurations assign the model an expert role and instruct it to determine whether a post is antisemitic. To account for ambiguity and context sensitivity, models may respond with \textit{Yes}, \textit{No}, or \textit{Unsure} \citep{goyal-etal-2025-momoe}.
Models are further instructed to consider speaker intent; in particular, quoting or reporting antisemitic content for purposes of criticism or condemnation is not to be labeled as antisemitic. Prompts returning explanations instruct the model to reference relevant sections of the provided conceptual representation (see Appendix~\ref{app:prompts} for the complete prompt structure).

The four prompt configurations share the same task formulation and differ only in the conceptual representation; BASE, which provides no external representation, therefore returns only a classification decision.  
IHRA provides a coarse-grained definition with explanatory and illustrative text, whereas STRUCT and STRUCT+EX use a fine-grained taxonomy corresponding to the \textit{Lexicon} chapters. 
Both conceptual representations are segmented into explicit \texttt{[Section X]}-style markers to facilitate  explicit concept referencing. STRUCT+EX extends STRUCT  with few-shot examples, using the first available  implicit antisemitic, explicit antisemitic, and non-antisemitic example for each concept from \textit{Lexicon} to ensure a reproducible selection procedure. 
We additionally quantified the overlap of the examples in STRUCT+EX with the Decoding dataset. Only 19 instances were found, all of which were antisemitic (0.6\% of all antisemitic instances). As an additional robustness check, we evaluate performance on this subset separately.

While STRUCT+EX is substantially longer, IHRA and STRUCT remain of comparable prompt length (see  Table~\ref{tab:prompt_configs}). However, even the longest prompt occupies only a small fraction of the available context windows of the evaluated models. 
The key distinction between the two representations therefore lies in their conceptual granularity and representational form:  IHRA embeds coarser categories in explanatory and discursive text, while STRUCT is a much more fine-grained but highly compressed list of concepts.

\begin{table}[h]
\centering
\caption{Prompt configurations used in the experiments.}
\label{tab:prompt_configs}
\small
\setlength{\tabcolsep}{2pt}
\begin{tabular}{lcccc}
\toprule
\textbf{Config.} & \textbf{Repr.} & \textbf{\# Concepts} & \textbf{Few-shot} & \textbf{\# Chars.} \\
\midrule
BASE      & None       & -- & --  & 491 \\
IHRA      & Definition & 12 & No  & 3,699 \\
STRUCT    & Taxonomy   & 41 & No  & 2,073 \\
STRUCT+EX & Taxonomy   & 41 & Yes & 18,605 \\
\bottomrule
\end{tabular}
\end{table}

We include all antisemitic posts and randomly sample 1{,}000 non-antisemitic posts per dataset. This yields stable estimates while keeping the computational cost manageable given the class-imbalance and the large number of model–prompt combinations evaluated. 

Referenced sections are extracted during post-processing and mapped onto the content groups (Table~\ref{tab:content_groups} in Appendix~\ref{app:annotation_schema_mapping}).
All metrics are computed on binary labels. Since manual inspection indicated that \textit{Unsure} responses typically reflected insufficient context for a positive classification, we treat them as \textit{No}.
Precision and F$_1$ are estimated via bootstrap resampling and inverse-probability weighting.

Results are reported on the balanced (i.e., equally weighted) union of both datasets and, where relevant, on the individual datasets. Statistical significance is assessed using $\chi^2$ tests with Holm correction; effect sizes are reported using Cohen's $h$.

\section{Results}

\subsection{Effect of External Conceptual Representation}\label{sec:res_rq1}

\begin{figure}[h]
    \centering
    \includegraphics[width=0.99\linewidth]{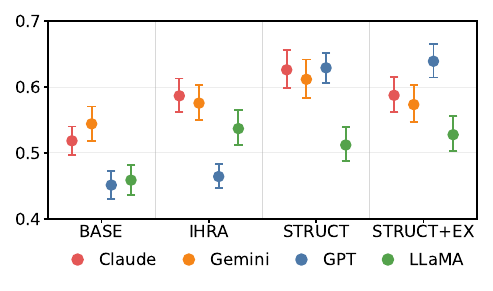}
\caption{F$_1$ scores with confidence intervals for each model and configuration on the balanced dataset union.}   
\label{fig:f1}
\end{figure}

Figure \ref{fig:f1} shows the F$_1$ scores of all models and configurations on the dataset union. Detailed results for all metrics and datasets are reported in Table \ref{tab:rq1_test_results} in Appendix~\ref{app:analysis_details}.

The three commercial models achieve similar peak F$_1$ scores (0.61--0.64), whereas LLaMA performs substantially worse (0.54). All models improve their F$_1$ scores when provided with external conceptual representations. Except for LLaMA, all models achieve their highest F$_1$ scores under STRUCT or STRUCT+EX. While LLaMA performs best under IHRA, its results for STRUCT and STRUCT+EX are comparable. 

Optimal settings differ between precision and recall. STRUCT and STRUCT+EX consistently yield the highest recall, often with substantial gains; for GPT and LLaMA, recall nearly doubles relative to BASE. The improvement in recall is significant even when comparing  STRUCT with IHRA, for all four models, with large effect sizes for Gemini. 
However, these gains generally come at the cost of lower precision, with GPT constituting the only exception. GPT is also the only model whose recall further improves through the inclusion of explicit examples. Because STRUCT+EX uses examples from \textit{Lexicon}, we evaluated the 19 overlapping Decoding instances. On this subset, STRUCT+EX performs at least as well as STRUCT for all models, indicating a local overlap effect. However, given the very small size of the subset, this effect cannot materially account for the overall performance differences between the two configurations. 
Moreover, GPT's precision remains consistently higher than recall across all configurations, while the remaining models achieve higher recall than precision in at least their best-performing setting. GPT also obtains the overall lowest recall.

Overall, performance trends are broadly consistent across datasets despite moderate differences in difficulty. 
Recall is consistently higher on Bloomington than on Decoding across all models and configurations, suggesting that Decoding is the more challenging dataset for the evaluated models. However, performance improvements associated with external representations, particularly STRUCT and STRUCT+EX, are substantially larger on Decoding than on Bloomington. Gemini largely closes the recall gap between datasets under these settings. Precision differences between datasets are less consistent. 
Consequently, F$_1$ scores remain uniformly higher on Bloomington across configurations.

Finally, all models  use the abstention option \textit{Unsure} (that was mapped to \textit{No} for metric computation) only rarely (typically $<=6\%$), particularly on Bloomington data and positive samples from Decoding. However, there are notable exceptions for negative Decoding samples. Gemini and GPT abstain considerably more often ($6-24 \%$ and $5-19 \%$ of cases, respectively), but interestingly, external conceptual representations reduce abstention for Gemini but increase it for GPT.

\subsection{Detection Performance Across Forms of Antisemitism}
\label{sec:res_rq2}

\begin{figure*}
    \centering
    \includegraphics[width=0.99\linewidth]{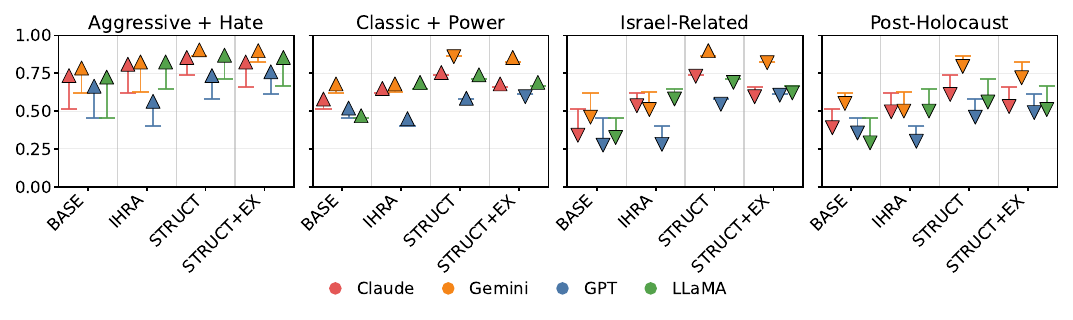}
    \caption{Recall for each content group. Arrows indicate gain/loss relative to average recall.}
    \label{fig:f1_AS}
\end{figure*}

We focus exclusively on posts labeled as antisemitic by human annotators to examine whether models differ in their ability to detect particular forms of antisemitism. 
Note that for multi-label posts, we treat every content group present in the human annotations as correct.
Figure~\ref{fig:f1_AS} shows the deviation of each content-group recall from the mean recall of the respective model and configuration; exact values and significance test results are reported in Table~\ref{tab:rq2_test_results} in Appendix~\ref{app:analysis_details}.

Across all models and configurations, aggressive antisemitic speech acts are detected most reliably. As reflected by the consistently positive deviations in Figure~\ref{fig:f1_AS}, these forms appear comparatively easy to identify, presumably because they often contain explicit derogatory language resembling overt hate speech. By contrast, Israel-related and post-Holocaust antisemitism constitute the most difficult content groups. In the BASE setting, Israel-related antisemitism typically exhibits the lowest recall, whereas post-Holocaust antisemitism remains difficult across all configurations.

Israel-related antisemitism exhibits the strongest dependence on prompting configuration. While recall is consistently low in BASE, STRUCT substantially improves performance across all models, at least doubling recall relative to BASE. STRUCT+EX yields similar or slightly smaller gains for most models, except for GPT, where explicit examples further increase recall.

The impact of IHRA is considerably less consistent. For GPT, IHRA preserves or decreases recall across all content groups. Claude and LLaMA generally benefit from IHRA, whereas Gemini exhibits mixed effects depending on the content group. Overall, STRUCT and STRUCT+EX produce the most robust improvements across models and forms of antisemitism. Most gains are statistically significant and frequently associated with large effect sizes. Even when compared with IHRA, STRUCT yields statistically significant improvements for almost all combinations of model and content group.

The dataset-level trends observed in Section~\ref{sec:res_rq1} also hold at the content-group level. Under BASE, recall on Decoding is generally lower than on Bloomington. However, STRUCT and STRUCT+EX reduce this gap considerably, and for some content groups, particularly Israel-related antisemitism, performance on Decoding approaches or even exceeds Bloomington.

Finally, external conceptual representations reduce performance disparities between content groups. In most cases, the standard deviation of content-group recalls decreases under STRUCT or STRUCT+EX, indicating more balanced detection performance across forms of antisemitism. These gains are primarily driven by strong improvements for previously underperforming categories, especially Israel-related antisemitism. Gemini exhibits the lowest overall variance across content groups.

\subsection{Grounded Explanations and Interpretability}\label{sec:res_rq3}

We next analyze how models reference conceptual categories in their explanations. For IHRA, we extract references to \textit{IHRA} sections; for STRUCT and STRUCT+EX, we extract references to \textit{Lexicon} chapters and evaluate the resulting predictions as a multi-label classification task.

\subsubsection{Number of References in Explanations}

Across all models and datasets, generated explanations reference substantially more concepts  than human annotations. This effect is particularly meaningful for Decoding, where annotators could assign an unrestricted number of labels, yielding $\sim\!\! 1.5$ labels per antisemitic post on average. In comparison, STRUCT produces  $2.5-3.5$ referenced chapters, roughly twice as many as human annotators.
The effect is substantially weaker for IHRA, where explanations typically reference $1.4-2.1$  sections per post. The number of  referenced concepts therefore scales strongly with the granularity of the conceptual representation, suggesting that fine-grained taxonomies encourage models to decompose antisemitic content into a larger number of simultaneously activated concepts. At the same time, STRUCT+EX consistently reduces the number of generated references relative to STRUCT across all models and datasets, indicating that explicit examples encourage more selective category assignment. Model-specific differences are also pronounced: Claude produces the largest number of references, followed by GPT and LLaMA, whereas Gemini remains substantially more selective.

The overproduction of referenced concepts persists even after aggregating codes into content groups. Human annotations contain $1.24 \pm 0.02$ content groups on average, compared with $1.64\pm 0.19$ in model references, corresponding to 1.3 times more referenced content groups on average. For details, see  Table~\ref{tab:rq3_distribution} in Appendix~\ref{app:analysis_details}.

\subsubsection{Distribution of Referenced Concepts}

\begin{figure*}
    \centering
    \includegraphics[width=0.99\linewidth]{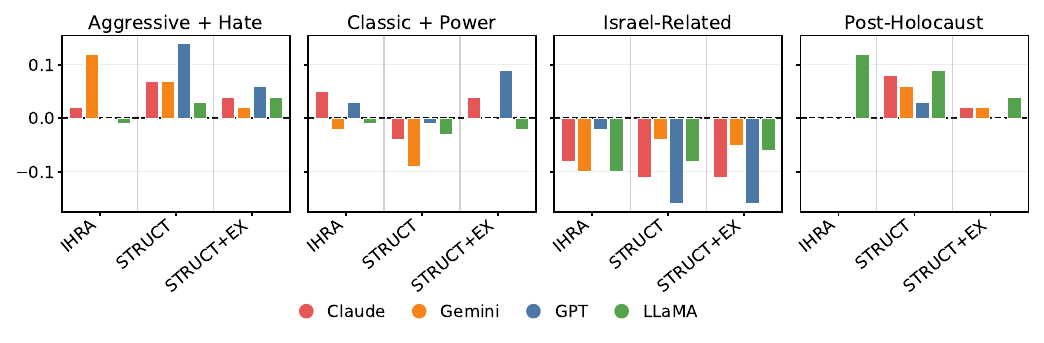}
        \caption{Deviation of model explanation distributions from human annotator distributions across content groups and configurations on the balanced dataset union.}
    \label{fig:rq3}
\end{figure*}

Figure~\ref{fig:rq3} shows the deviation between the distribution of concepts referenced in model explanations and the distribution in human annotations.

Across models and configurations, aggressive speech acts are consistently over-represented in model explanations, whereas Israel-related concepts are systematically under-referenced. This pattern is consistent with the detection results from the previous section: models both detect aggressive antisemitic content more reliably and invoke the corresponding concepts more frequently, whereas Israel-related antisemitism exhibits both substantially lower recall and fewer conceptual references, especially under BASE and IHRA.
Post-Holocaust antisemitism exhibits a contrasting pattern. Despite comparatively low recall across models, post-Holocaust concepts are frequently over-referenced under STRUCT and, to a lesser extent, STRUCT+EX.

\subsubsection{Classification Performance via References}

Across all models, configurations and datasets, Israel-related antisemitism is classified most accurately, with F$_1$ scores between 0.79 and 0.94. This is notable given that Israel-related content was substantially more difficult to detect as antisemitic, especially under BASE and IHRA. Once identified as antisemitic, however, models appear to assign Israel-related concepts comparatively reliably.
By contrast, aggressive and hateful speech acts, while comparatively easy to detect as antisemitic, are classified considerably less precisely, with F$_1$ scores between 0.20 and 0.56, mainly caused by very low precision results (see Table~\ref{tab:rq3_distribution}). Post-Holocaust antisemitism remains the most challenging category overall, both for detection and reference-based classification, with consistently low performance across models and configurations.

These results should be interpreted in light of the systematic overproduction of references: precision is systematically reduced because models cite substantially more concepts than annotators. Accordingly, configurations with high over-referencing are associated with lower precision values, resulting in every third referenced content group being false. Detailed precision, recall and F$_1$ scores by content group, dataset, model and configuration are reported in Table~\ref{tab:rq3_distribution} in Appendix~\ref{app:analysis_details}.

\subsection{Large-Context Provision}\label{sec:res_rq4}

So far, our results show that prompting with compact fine-grained conceptual representations is associated with substantially better antisemitism detection and reference-based classification. This raises the question whether providing a substantially richer  resource comprising theoretical background, historical context, and extensive examples would further improve model behavior. To investigate this, we leverage the large-context capabilities of Gemini by providing the complete \textit{Lexicon} instead of the compact taxonomy alone.

Contrary to expectations,  LEXICON  does not improve quantitative performance (see Appendix~\ref{app:analysis_details}. Detection and classification metrics remain comparable to STRUCT and STRUCT+EX despite the notably larger amount of contextual information available to the model. 

\subsubsection{Qualitative Error Analysis and Manual Explanation Inspection}

We applied inductive category formation following \citet{mayring2014} to analyze all model errors, that is explanations for all false positives and negatives. We additionally inspected 450 generated explanations of correct predictions. 

Many false positives involve posts that quote, criticize, or sarcastically reference antisemitic statements. In several cases, the model even correctly identifies sarcasm or condemnation within its explanation.
Explanations of false positives also frequently indicate strong reliance on salient lexical cues such as \textit{kikes} or \textit{Holocaust}.  
Another group of false positives is better explained by differences between annotation frameworks than by model failure. For example, Holocaust distortion is explicitly operationalized in the \textit{Lexicon} but not in the \textit{IHRA}, which serves as the annotation basis for Bloomington.  
We further observe cases of model overconfidence, as the respective post showed a certain level of missing context. In some cases, the model even explicitly acknowledged respective uncertainty in the explanation.

False negatives predominantly involve subtle, indirect, or justificatory forms of antisemitism. These include Holocaust denial framed through legalistic reasoning (e.g., arguing that deaths due to disease in concentration camps do not constitute murder due to alleged lack of intent), indirect endorsements of violence against the Jewish state, symbolic delegitimization such as quotation marks around ``Israel,'' and stereotypical allegations against Jews framed as accusations against Israel as a Jewish collective. Other missed cases include posts in which antisemitic narratives are framed as personal or political opinion.
When posts rely on misinformation, historical distortions, or conspiratorial assumptions, explanations frequently accept these premises at face value rather than contextualizing or challenging them. This is less pronounced for explicit Holocaust denial, but remains visible for more subtle or historically less established antisemitic narratives.

Overall, the model generally identifies relevant antisemitic concepts and often handles implicit linguistic strategies such as intentional misspellings (e.g., ``Juice'' for ``Jews'') or dog whistles. However, when multiple tropes occur simultaneously, explanations typically focus on the most salient cue. In the presence of explicit slurs, explanations concentrate almost exclusively on lexical markers even when other antisemitic tropes are more central to the post. References to multiple chapters often reflect elaborations of a single detected pattern rather than recognition of conceptually distinct tropes.

\section{Discussion and Future Work}

Our results show that \textbf{external conceptual representations are consistently associated with  improved antisemitism detection and classification}, confirming \citet[p. 56]{becker_etal_report6_2024} and contrasting prior findings on hate speech \citep{melis-etal-2025-modular,roy-etal-2023}. While \textit{IHRA} generally improves performance, the gains remain comparatively modest, similar to \citet{patel-etal-2025}. In contrast, the fine-grained taxonomic representations consistently achieve the largest recall improvements across datasets and models. Particularly striking is the observation that providing a compact list of antisemitic concepts without further elaboration often performs best, whereas supplying the complete \textit{Lexicon} does not yield additional improvements. 

One possible explanation is that the compact taxonomy provides more effective conceptual cues than the more discursive \textit{IHRA} definition. Rather than embedding concepts within explanatory and normative text intended for human interpretation, STRUCT presents them as a fine-grained, explicitly organized inventory that may facilitate retrieval of relevant concepts during inference. Because the models are simultaneously required to justify their decisions through free-text explanations, the taxonomy may therefore guide the selection of concepts while also relying on knowledge already encoded in the model. Moreover, the two representations differ not only in granularity but also in conceptual coverage. More than half of the taxonomy concepts have no sufficiently explicit counterparts in \textit{IHRA}; for example, the ``affirmation of Nazis, Hitler and the Holocaust'' is represented explicitly in the taxonomy, whereas under \textit{IHRA} it must be inferred from broader concepts such as ``justifying the killing or harming of Jews in the name of a radical ideology'' or ``hatred toward Jews'', which are both logical consequences of Nazism but require additional reasoning steps that may make retrieval of, and the decision for, the appropriate concept more difficult. 
However, the present experiments do not disentangle this potential interpretation from alternative explanations. A larger number of concepts may simply increase the model's sensitivity to positive predictions, consistent with the observed precision-recall trade-off, and other prompt characteristics may likewise contribute. Establishing causal mechanisms would require additional controls, including varying taxonomy granularity, taxonomy manipulations, example-only prompting, retrieval-based prompting using the \textit{Lexicon}, or alternative prompt structures, while controlling for prompt length. 

At the same time, the improved recall using a fine-grained taxonomy comes at the cost of reduced precision. Neither curated examples nor the substantially richer \textit{Lexicon} mitigated this \textbf{precision-recall trade-off}. This is consistent with the hypothesis that additional contextual detail alone may be insufficient to improve conceptual selectivity once the relevant concepts have already been activated. 

\textbf{Performance differs substantially between the two datasets}. Across all models, Bloomington achieves considerably higher recall under BASE, indicating that it is the easier benchmark in our setting. This is consistent with the different construction of the datasets: Bloomington consists of Twitter posts sampled using (explicit) keywords, whereas Decoding comprises discourse-sampled comments from mainstream media, thus presumably containing more implicit forms of antisemitism. Against this background, STRUCT yields substantially larger improvements on Decoding than on Bloomington. One possible explanation is that fine-grained conceptual representations provide greater benefit on more challenging detection tasks. However, an alternative explanation is that STRUCT is more closely aligned with the Decoding annotation framework, whereas Bloomington was annotated according to \textit{IHRA}. This interpretation alone, however, is insufficient, as STRUCT consistently outperforms \textit{IHRA} even on Bloomington. Disentangling the contributions of dataset difficulty and annotation-framework alignment therefore requires evaluation on additional independently annotated datasets.

Our findings further reveal \textbf{systematic differences between forms of antisemitism}. Across all models and prompting configurations, aggressive speech is detected most reliably, aligning with prior observations that explicit hate is easier for LLMs to recognize \cite{ocampo-etal-2023-depth,elsherief-etal-2021-latent}.
By contrast, Israel-related and post-Holocaust antisemitism are substantially more challenging, confirming existing research \citep{adl_report_2025}. 
However, the recall for \textbf{Israel-related antisemitism} doubles for all models, with the fine-granular taxonomy, while the improvement is modest with \textit{IHRA}. This is interesting given ongoing objections that \textit{IHRA} leads to a general interpretation of criticism of Israel as antisemitic. At the same time, Israel-related antisemitism is the most correctly classified content group, across models and configurations, once detected. A plausible explanation is the presence of lexical cues such as \textit{Israel}.

\textbf{Post-Holocaust antisemitism} remains consistently difficult both for detection and classification. At the same time, the fine-grained taxonomy substantially increases references to post-Holocaust concepts in explanations despite comparatively low recall for the corresponding posts. The overly broad conceptual activation is constrained when explicit examples are provided. These findings point toward difficulties in calibrating when historically contextualized narratives should be interpreted as antisemitic. Interestingly, LLaMA exhibits similar overproduction even under \textit{IHRA}, potentially reflecting model-specific differences in how antisemitism concepts are internally structured. One possible explanation is that several forms of post-Holocaust antisemitism, such as reversal of victims and perpetrators or resentment toward Holocaust remembrance, are particularly rooted in German post-war discourse and may therefore be less strongly represented in predominantly English-language training data. Future work should investigate these patterns more directly, including cross-lingual analyses and targeted evaluation on historically contextualized antisemitic narratives.

A qualitative analysis of Lexicon-prompted Gemini model errors confirmed previous findings on prompting-based detection and classification of hate speech, discussed in Related Work.
We observed ongoing difficulties with \textbf{sarcasm, quotation, and condemnation} despite explicit prompt instructions to account for speaker intent. In line with \citet{tang-etal-2023-large}, we additionally observed \textbf{laziness and reliance on lexical cues},  suggesting surface-level conceptual matching rather than robust contextual reasoning. When multiple antisemitic patterns co-occur, explanations typically elaborate only the most salient or easily detectable one.
Another type of error is \textbf{model overconfidence}, reflected in the underuse of the abstention option. Across most model and configuration settings, the abstention label (\textit{Unsure}) was selected in only a small percentage of cases, consistent with prior work \citep{madhusudhan-etal-2025-llms,wen-etal-2024-characterizing}. At the same time, our qualitative analysis of the explanations revealed that models often expressed uncertainty while still producing a binary classification. The variation in abstention behavior across specific conditions and datasets, as well as the discrepancy between explicit abstention and uncertainty expressed in explanations, represents an important direction for future research.

Generally Gemini explanations with Lexicon showed plausible reasoning, allowing indirect classification of antisemitic forms and providing insight into which concepts models attend to. 
However, we observe ignorance of antisemitism in posts relying on \textbf{misleading or historically distorted premises}. In such cases, explanations often accept problematic assumptions at face value rather than challenging them. 
This reveals a broader limitation of LLMs for social science research: the absence of contextual correction or epistemic stance-taking.

\section{Limitations}

To enable comparability across datasets, we mapped Decoding Antisemitism annotation codes to both \textit{Lexicon} chapters and \textit{IHRA} sections. While this mapping was conducted carefully, some ambiguity persists at the level of individual sections. Also, a few Decoding codes fell under two \textit{IHRA} sections simultaneously, which meant an inflation of the number of referenced codes. Aggregating at the level of content groups mitigates this issue but necessarily reduces analytical granularity. Consequently, our analyses operate at the content-group level rather than the level of individual \textit{Lexicon} chapters or \textit{IHRA} sections, limiting insights into finer-grained distinctions between specific antisemitic tropes.

A second limitation concerns the operationalization of external conceptual resources. 
Our large-context experiment evaluates direct in-context prompting of the complete \textit{Lexicon}, rather than the intrinsic utility of the resource itself. Consequently, the absence of performance gains does not imply that the additional conceptual information is uninformative, but may instead reflect limitations of long-context prompting in retrieving and utilizing relevant information.
Questions of how fine-grained and up-to-date conceptual knowledge can be maintained and integrated efficiently remain largely unresolved. Possible directions include retrieval-based prompting (e.g., RAG) and more structured representations such as knowledge graphs \citep{yang_comprehensive_2025}. However, such approaches remain underdeveloped for antisemitism research, with  rather small Wikipedia-based knowledge bases utilized so far \citep{halevy_2023}. Future work should therefore investigate more systematic forms of structured conceptual grounding and their interaction with contextual reasoning.

Another limitation concerns the provenance of the \textit{Lexicon} used in the additional large-context experiment (Section~\ref{sec:res_rq4}). Many illustrative examples originate from the Decoding dataset used in our experiments, potentially advantaging Lexicon-based prompting. While these examples have also appeared in published reports and may therefore form part of the models' broader pretraining knowledge, this overlap cannot be fully disentangled. Importantly, however, Lexicon-based prompting also achieves strong recall on the independently annotated  Bloomington dataset, suggesting that its benefits are not solely confined to the source corpus. We additionally observe no systematic overrepresentation of early or late \textit{Lexicon} chapters in generated references, indicating no obvious positional attention bias in the long-context setting. However, because the table of contents was part of the provided \textit{Lexicon},  future work should investigate attention allocation in long conceptual documents more directly.

Finally, while evaluating four models permits initial assessment of which findings generalize across systems and which remain model-specific, broader model coverage would naturally yield richer comparative insights. Our selection of four predominantly commercial models reflects a deliberate balance between contrastive breadth and interpretive manageability. Extending the analysis to more diverse architectures, scales, and open-weight systems remains an important direction for future work.

\section{Ethical Considerations}
Antisemitism detection is a high-stakes task in which both false negatives and false positives may have substantial societal consequences, though with different impacts. Our results show that different conceptual representations have a strong impact on the precision-recall trade-off and affect the detection of particular forms of antisemitism. These findings highlight the importance of carefully evaluating prompting strategies before considering their use in moderation settings. Moreover, our analysis reveals systematic limitations of current LLMs, including overconfidence, reliance on lexical cues, and reduced performance on specific forms of antisemitism. Generated explanations should therefore only be used with appropriate human oversight.

Our experiments use expert-annotated datasets derived from publicly available online sources and commercial LLM APIs. The prompts ask models to identify rather than generate or endorse antisemitic content. According to the providers' API policies, inputs are not used for model training, reducing concerns about exposing models to harmful content through our experiments.

\bibliography{references}

\appendix
\section{Appendix}

\subsection{Antisemitism}
\label{app:antisemitism}

Adorno describes antisemitism as ``the rumor about the Jews'' \citep{adorno1999minima}, stressing that antisemitism is not explainable by what Jews do, but what the antisemite believes them to do.
Reflecting its persistence as a ``cultural constant'' \cite{schwarz-friesel_judenhass_2019} over more than 2,000 years and its capacity to adapt to changing historical, social, and political contexts, antisemitism is often referred to as ``the longest hatred'' \citep{wistrich1994antisemitism}. 
The ideology manifests through a heterogeneous set of stereotypes and narratives targeting Jews both as individuals and as a collective. Many of these stereotypes have deep historical roots, yet they are repeatedly rearticulated in new forms, particularly, though not exclusively, in times of social or political crisis.

A distinctive feature of antisemitism is its inherently contradictory structure. Jews are frequently portrayed simultaneously as inferior and superior, weak and powerful, victimized and manipulative. This paradoxical imagery enables antisemitic discourse to accommodate a wide range of conspiracy narratives, in which Jews are imagined as acting covertly or exercising hidden influence. As Porat notes, antisemitism is particularly difficult to grasp ''since antipathy to Jews involves a deep-seated emotional dimension as well as a conglomerate of historic religious, political, and economic elements'' \citep[476]{porat2020}.

Historically, Christian anti-Judaism represents the oldest articulation of antisemitism \citep{wistrich1994antisemitism}. The term itself was made popular by Marr in the late 19th century, expressing the emergence of racial antisemitism which was no longer (primarily) religiously motivated, seeing Jews as a race rather than a religious community \citep{marcus2015}.
Under National Socialism, antisemitism reached its most extreme and industrialized form in the Holocaust.
After 1945, antisemitic discourse did not disappear but instead adapted to new normative environments, giving rise to forms commonly described as post-Holocaust antisemitism.
These include Holocaust denial and distortion, inversion of victim–perpetrator roles, and Israel-related antisemitism that targets the Jewish state as a Jewish collective \citep{marcus2015}.

The longevity of antisemitism has resulted in a range of linguistic and narrative patterns. Contemporary antisemitism is not limited to explicit Jew hatred but is frequently articulated in encoded and implicit ways. Prior research shows that antisemitism, offline and online, can be expressed in a variety of coded forms, including dog whistles, allusions, metaphors, wordplay, and other semiotic markers \citep{schwarzfriesel2020}. At the same time, antisemitism may also be expressed  through conspiracy narratives that attribute hidden power, malevolent intent, or global control to abstract entities such as `elites' without mentioning Jews directly \cite{finkelstein_antisemitic_2020,comerford2021}. 

Common forms of antisemitism are defined in the IHRA working definition of antisemitism \cite{ihra_2016}, which consists of a general paragraph accompanied by eleven examples of common contemporary manifestations.  
The definition is the result of decade-long debates involving multiple stakeholders such as scholars and policy makers together with representatives of the Jewish community, and is nowadays adopted by the majority of Jewish organizations as well as 35 of the IHRA member states. The indeed interesting history of articulating the necessity for a definition and the process for formulating one can be found in \citet{marcus2015,porat2020}.

\subsection{Dataset details}\label{app:dataset_details}
While both datasets are grounded in the \textit{IHRA} definition, they differ along several key dimensions: Bloomington relies on keyword-based sampling and a coarse-grained, predominantly single-label annotation scheme, whereas Decoding uses discourse-driven sampling and a fine-grained, multi-label annotation scheme.

\subsubsection{Bloomington}
\label{sec:data_bloomington}
The Bloomington dataset\footnote{The binary labeled dataset is available on Zenodo: https://zenodo.org/records/14448399; CC BY 4.0 license. The section labels were made available upon request.} consists of English-language posts from Twitter (now X) \citep{jikeli_etal_2021} containing one of the keywords \textit{jews}, \textit{israel}, \textit{kikes}\footnote{The slur \textit{kikes} has a long history in antisemitic discourse. In the \textit{Studies in Prejudice}, Adorno et al.\ describe its use in constructing ``the standard division of Jews into two groups, the good ones and the bad ones, a division frequently expressed in terms of the `white' Jews and the `kikes''' \citep{adorno1950}.}, and \textit{zionazi*}.
Posts were annotated by two field experts following \textit{IHRA} in its twelve-section formulation. For posts labeled as antisemitic, annotators selected the most dominant \textit{IHRA} section as a rationale. Disagreements regarding antisemitism labels were resolved through discussion, whereas disagreements on section assignment were retained, allowing antisemitic posts to be associated with at most two section labels \cite{jikeli2024}. 

The dataset contains 11{,}311 posts, of which 17.3\% are labeled as antisemitic. Prevalence varies substantially across keyword-based subsamples: posts containing \textit{zionazi} are predominantly antisemitic ($88.3\%,  n=529$), followed by \textit{kikes} ($34.3\%, n=283$); in contrast, posts containing \textit{jews} ($16.1\%, n=6{,}395$) or \textit{israel} ($11.4\%, n=4{,}134$) are mostly non-antisemitic.

\subsubsection{Decoding}

The Decoding dataset\footnote{The dataset has been provided by the PI of the Decoding Antisemitism research project upon request.} comprises comment sections of a wide range of mainstream news outlets and social media platforms, with threads selected around public events and topics expected to trigger antisemitic discourse \citep[p.~544]{decoding_lexicon_2024}. For the present study, we use the English-language subset of the corpus, comprising 42{,}799 comments, most of which originate from Facebook (52.8\%), followed by YouTube (21.1\%), mainstream news outlets such as The Guardian and the BBC (18\%), and Twitter (8.1\%).

Considering  \textit{IHRA} to be insufficiently precise for systematic corpus analysis of contemporary antisemitism \citep[p.~11]{decoding_lexicon_2024}, the definition was extended and refined into an annotation scheme comprising 46  codes for antisemitic content. The resulting scheme specifies particular forms of post-Holocaust antisemitism such as the tabooing of criticism \citep[p.~309]{decoding_lexicon_2024}; differentiates stereotypes subsumed under broader \textit{IHRA} categories, including lie and deceit \citep[p.~93]{decoding_lexicon_2024} and vengefulness \citep[p.~107]{decoding_lexicon_2024}; and explicitly includes Israel-related narratives such as generalized apartheid analogies and calls for boycott, divestment, and sanctions (BDS). In addition, distinctions between antisemitism targeting Jewish individuals and the Jewish state are reorganized compared to \textit{IHRA}.
The annotators were allowed to assign an unlimited number of these codes to antisemitic comments. 

Overall, 7.9\% of comments are labeled as antisemitic, substantially fewer than in the Bloomington dataset. This difference likely reflects the predominance of moderated mainstream platforms as well as the discourse-driven sampling strategy, which does not rely on explicit keywords related to Jewish life or Israel.

\begin{table*}[h]
\centering
\caption{
Joint content groups used for analysis, including definitions and their alignment with knowledge bases.
}
\label{tab:content_groups}
\small
\setlength{\tabcolsep}{6pt}
\begin{tabular}{p{2.5cm} p{8.7cm} p{1.2cm} p{1.3cm}}
\toprule
\textbf{Content group} 
& \textbf{Description}
& \textbf{Lexicon chapters}
& \textbf{IHRA sections} \\
\midrule

Aggressive+Hate
& Explicit expressions of hostility toward Jews, including hate speech, aggressive language, and calls for violence.
& 37--42
& 0, 1 \\

Classic+Power
& Stereotypical allegations against Jews, particularly narratives of power, manipulation, conspiracy, or malevolent influence.
& 2--15
& 2, 3, 6 \\

Israel-Related
& Antisemitic expressions targeting Israel as the Jewish state, including collective attributions, delegitimization, and analogies with Nazism.
& 28--36
& 7, 8, 9, 10 \\

Post-Holocaust
& Forms of antisemitism emerging after 1945, including denial, relativization, inversion, and secondary antisemitic narratives.
& 16--27
& 4, 5, 11 \\

\bottomrule
\end{tabular}
\end{table*}

\subsubsection{Annotation Frameworks and Joint Representation}
\label{app:annotation_schema_mapping}

Given that they are both used empirically in established research projects, and that \textit{Lexicon} is based on \textit{IHRA}, both annotation schemes show little intersection: The majority of the concepts presented in \textit{Lexicon} can neither implicitly nor explicitly be found in \textit{IHRA}. On the other hand, \textit{IHRA} elaborates concepts much more intensely, resulting in a longer text. To enable joint analysis, we  developed a mapping that aligns the fine-grained codes of Decoding with both \textit{Lexicon} chapters and \textit{IHRA} sections. When a single Decoding code corresponds to multiple \textit{IHRA} sections, all relevant sections are retained; such instances therefore contribute to multiple \textit{IHRA} categories. 

We further abstract both annotation schemes into a shared set of higher-level \textit{content groups} that capture recurring dimensions of antisemitic discourse and enable robust comparison across datasets and conceptual representations. Table~\ref{tab:content_groups} summarizes the four content groups, together with their definitions and the corresponding \textit{IHRA} sections and \textit{Lexicon} chapters.

While Israel-related antisemitism can be understood historically as a subset of post-Holocaust antisemitism (see Appendix~\ref{app:antisemitism}), we treat it as a separate content group for analytical reasons: In both \textit{IHRA} and \textit{Lexicon}, Israel-related antisemitism is elaborated in multiple sections or chapters;  empirically it constitutes a substantial proportion of antisemitic content in both datasets. 

\subsection{Preprocessing}\label{app:preprocessing}

Both datasets were preprocessed using the same pipeline. We removed URLs, removed explicit markers indicating reply structure and stripped HTML markup while preserving the underlying textual content. User handles occurring at the beginning of a post were removed, while those appearing within the text were retained, as they may function as meaningful references to individuals or organizations (e.g., \textrm{@netanyahu}, \textrm{@WashingtonPost}). 
In almost all cases, models produced grounded explanations for a \textit{Yes} classification, meaning they provided section and chapter references. Only for three posts, LLaMA did not mention a section number, but referenced the section by quoting it. We manually inspected these explanations and recoded them to the section number the model was quoting.

\newpage
\subsection{Prompts}\label{app:prompts}

The modular prompt template shared by all conditions, is shown in the box below.
The \texttt{system\_prompt}, shared task instructions, and JSON
output format are identical across all conditions.

\noindent
\textbf{Conceptual representation component variants:}

\noindent
\textbf{NO KB}: None;

\noindent
\textbf{IHRA}: IHRA Working Definition sections 0--11;

\noindent
\textbf{STRUCT}: taxonomy labels 2--42;

\noindent
\textbf{STRUCT+EX}: taxonomy labels 2--42 + examples;

\noindent
\textbf{LEXICON}: Decoding Antisemitism Lexicon (full document as PDF).

\begin{mdframed}\label{box:prompts}
{\small
\noindent\textbf{[System Prompt]}

\medskip\noindent
You are an expert in classifying social media posts with regard
to antisemitic content.

\medskip\noindent
\textbf{[Configuration task instruction]}

\medskip\noindent
Use the following definition for your classification decision: \{IHRA\}
|
\noindent
Use the following taxonomy of antisemitic concepts for your classification decision: \{STRUCT\}
|
\noindent
Use the following taxonomy of antisemitic concepts for your classification decision. In most cases, antisemitic and non-antisemitic examples are provided to guide your decision. \{STRUCT+EX\}

\medskip\noindent
\textbf{[Shared task instructions]}

\medskip\noindent
\textbf{Decide} whether the post below contains antisemitic speech. Thereby consider the \textbf{intent} of the message.
In particular, \textbf{quoting or reporting} antisemitism for
the purpose of criticism or condemnation should NOT be labeled
as antisemitism.
Answer preferably using `Yes' or `No'.
If the decision is not possible, you can use `Unsure'.

\medskip
\noindent\textbf{[Output format (JSON conditions)]}

\begin{verbatim}
{
  "index": <original index of the text>,
  "classification": "<Yes | No | Unsure>",
  "explanation": "<brief explanation with 
  reference to relevant taxonomy categories 
  by number and title>"
}
\end{verbatim}

\noindent Only output a JSON array of objects, no extra text.
}
\end{mdframed}

\subsection{Analysis details}\label{app:analysis_details}

\begin{table*}[t]
\centering
\small
\setlength{\tabcolsep}{6pt}
\caption{
Model performance across all models on Bloomington (B) and Decoding (D) and their balanced union (U). Best performing configuration  per model, metric and dataset is highlighted in bold. Significance is tested with BASE as reference group. Statistically significant differences are indicated by $\uparrow$ or $\downarrow$; large effects (Cohen's $\vert h\vert \geq 0.5$) are marked with $^\ast$.}
\label{tab:rq1_test_results}
\begin{tabular}{l l lll!{\hspace{13pt}}lll!{\hspace{13pt}}lll}
\toprule
\textbf{Model}
& \textbf{Cfg} 
& \multicolumn{3}{c}{\textbf{Precision}}
& \multicolumn{3}{c}{\textbf{Recall}}
& \multicolumn{3}{c}{\textbf{F1}} \\
\cmidrule(lr){3-11}
& 
& B & D & U
& B & D & U
& B & D & U\\
\specialrule{0.1em}{0.5em}{0.5em}  
\multirow{4}{*}{Claude}
& BASE
& \textbf{.65} & \textbf{.67} & \textbf{.66}
& .55 & .33 & .44
& .60 & .44 & .52\\
\cmidrule(l){2-11}
& IHRA
& .59 $\downarrow$ & .63 & .61
& .68 $\uparrow$ & .48 $\uparrow$ & .58 $\uparrow$
& .63 $\uparrow$ & .54 $\uparrow$ & .59\\
\cmidrule(l){2-11}
& STRUCT
& .56 $\downarrow$ & .54 $\downarrow$ & .55
& \textbf{.79 $\uparrow^{\ast}$} & \textbf{.67 $\uparrow^{\ast}$} & \textbf{.73 $\uparrow^{\ast}$}
& \textbf{.65 $\uparrow$} & \textbf{.60 $\uparrow$} & \textbf{.63 $\uparrow$}\\
\cmidrule(l){2-11}
& STRUCT+EX
& .58 $\downarrow$ & .53 $\downarrow$ & .56
& .71 $\uparrow$ & .54 $\uparrow$ & .63 $\uparrow$
& .64 $\uparrow$ & .54 $\uparrow$ & .59\\
\specialrule{0.1em}{0.5em}{0.5em}  
\multirow{5}{*}{Gemini}
& BASE
& .57  & .52 & .54
& .64  & .46 & .55
& .60  & .49 & .54\\
\cmidrule(l){2-11}
& IHRA
& \textbf{.61} $\uparrow$ & \textbf{.55} & \textbf{.58}
& .68 $\uparrow$ & .48 $\uparrow$ & .58 $\uparrow$
& \textbf{.64} $\uparrow$ & .51 & .58  \\
\cmidrule(l){2-11} 
& STRUCT
& .49 $\downarrow$ & .45 & .47
& \textbf{.88 $\uparrow^{\ast}$}   & \textbf{.87 $\uparrow^{\ast}$} & \textbf{.87 $\uparrow^{\ast}$} 
& .63  & \textbf{.59 $\uparrow$} & \textbf{.61} \\
\cmidrule(l){2-11} 
& STRUCT+EX
& .49 $\downarrow$ & .40 $\downarrow$ & .44 $\downarrow$
& .83 $\uparrow$   & .80 $\uparrow^{\ast}$ & .82 $\uparrow^{\ast}$
& .61  & .53 & .57\\
\cmidrule(l){2-11}
& LEXICON
& .47 $\downarrow$ & .45 & .46
& .84 $\uparrow$   & .85 $\uparrow^{\ast}$ & .85 $\uparrow^{\ast}$
& .60  & \textbf{.59 $\uparrow$} & .59\\
\specialrule{0.1em}{0.5em}{0.5em}  
\multirow{4}{*}{GPT}
& BASE
& .57  & .62 & .60
& .51  & .26 & .38
& .54  & .36 & .45\\
\cmidrule(l){2-11}
& IHRA
& \textbf{.74 $\uparrow$} & \textbf{.84} & \textbf{.79}
& .49 & .21 $\downarrow$ & .35
& .59 $\uparrow$ & .34 & .46   \\
\cmidrule(l){2-11}
& STRUCT
& .70 $\uparrow$ & .78 & .74
& \textbf{.64 $\uparrow$}   & .48 $\uparrow$ & .56 $\uparrow$
& \textbf{.67 $\uparrow$}  & .59 $\uparrow$ & .63 $\uparrow$ \\
\cmidrule(l){2-11}
& STRUCT+EX
& .68 $\uparrow$ & .70 & .69
& \textbf{.64 $\uparrow$}   & \textbf{.55 $\uparrow^{\ast}$} & \textbf{.60 $\uparrow$}
& .66 $\uparrow$  & \textbf{.61 $\uparrow^{\ast}$} & \textbf{.64 $\uparrow$}\\
\specialrule{0.1em}{0.5em}{0.5em}  
\multirow{4}{*}{LLaMA}
& BASE
& \textbf{.66} & \textbf{.48} & \textbf{.57}
& .48 & .29 & .38
& .56 & .36 & .46\\
\cmidrule(l){2-11}
& IHRA
& .51 $\downarrow$ & .44 & .48 
& .67 $\uparrow$ & .55 $\uparrow^{\ast}$ & .61 $\uparrow$
& \textbf{.58} & \textbf{.49 $\uparrow$} & \textbf{.54}\\
\cmidrule(l){2-11}
& STRUCT
& .48 $\downarrow$ & .33 $\downarrow$ & .41 $\downarrow$
& \textbf{.73 $\uparrow^{\ast}$} & \textbf{.66 $\uparrow^{\ast}$} & \textbf{.70 $\uparrow^{\ast}$}
& \textbf{.58} & .44 $\uparrow$ & .51\\
\cmidrule(l){2-11}
& STRUCT+EX
& .50 $\downarrow$ & .41 & .45 
& .68 $\uparrow$ & .59 $\uparrow^{\ast}$ & .64 $\uparrow^{\ast}$
& .57 & .48 $\uparrow$ & .53\\
\bottomrule
\end{tabular}
\end{table*}

\begin{table*}[t]
\centering
\small
\footnotesize
\setlength{\tabcolsep}{2pt}
\caption{Recall of antisemitism detection, disaggregated by model, configuration,  content group and dataset (B = Bloomington, D = Decoding, U = balanced union). Significance is tested with BASE as reference group. Statistically significant differences are indicated by $\uparrow$ or $\downarrow$; large effects (Cohen's $\vert h\vert \geq 0.5$) are marked with $^\ast$.}
\label{tab:rq2_test_results}
\begin{tabular}{l l lll!{\hspace{13pt}} lll !{\hspace{13pt}} lll !{\hspace{13pt}}lll}
\toprule
\textbf{Model}
& \textbf{Cfg}
& \multicolumn{3}{l}{\textbf{Aggressive+Hate}}
& \multicolumn{3}{l}{\textbf{Classic+Power}}
& \multicolumn{3}{l}{\textbf{Israel-related}}
& \multicolumn{3}{l}{\textbf{Post-Holocaust}} \\
\cmidrule(lr){3-14}
& & B & D & U & B & D & U & B & D & U & B & D & U \\
\specialrule{0.1em}{0.5em}{0.5em}
\multirow{4}{*}{Claude}
& BASE
& .82 & .65 & .74
& .78 & .38 & .58
& .41 & .27 & .34
& .49 & .29 & .39 \\
\cmidrule(l){2-14}
& IHRA
& .86 & .76 & .81 
& .83 & .47 $\uparrow$ & .65 $\uparrow$
& .58 $\uparrow$ & .49 $\uparrow$ & .54 $\uparrow$
& .62 & .37 $\uparrow$ & .50 $\uparrow$\\
\cmidrule(l){2-14}
& STRUCT
& .87 & .84 & .86 $\uparrow$
& .86 $\uparrow$ & .65 $\uparrow^{\ast}$ & .76 $\uparrow$
& .75 $\uparrow^{\ast}$ & .71 $\uparrow^{\ast}$ & .73 $\uparrow^{\ast}$
& .67 & .55 $\uparrow^{\ast}$ & .61 $\uparrow$\\
\cmidrule(l){2-14}
& STRUCT+EX
& .85 & .80 & .83
& .83 & .53 $\uparrow$ & .68 
& .64 $\uparrow$ & .55 $\uparrow^{\ast}$ & .60 $\uparrow^{\ast}$
& .59 & .47 $\uparrow$ & .53 $\uparrow$ \\
\specialrule{0.1em}{0.5em}{0.5em}
\multirow{4}{*}{Gemini}
& BASE
& .82 & .75 $\;\;$ & .79
& .85 & .51 & .68
& .51 & .41 & .46
& .71 & .39 & .55 \\
\cmidrule(l){2-14}
& IHRA
& .87 & .78 & .83
& .86 & .50 & .68
& .56 $\uparrow$ & .51 $\uparrow$ & .51 $\uparrow$
& .64 & .36 & .50 \\
\cmidrule(l){2-14}
& STRUCT
& .92 & .89 & .91 $\uparrow$
& .90 $\uparrow$ & .82 $\uparrow^{\ast}$ & .86 $\uparrow$
& .87 $\uparrow^{\ast}$ & .93 $\uparrow^{\ast}$ & .90 $\uparrow^{\ast}$
& .83 & .76 $\uparrow^{\ast}$ & .80 $\uparrow^{\ast}$ \\
\cmidrule(l){2-14}
& STRUCT+EX
& .91 & .89 & .90 $\uparrow$
& .91 $\uparrow$ & .80 $\uparrow^{\ast}$ & .86 $\uparrow$
& .79 $\uparrow^{\ast}$ & .85 $\uparrow^{\ast}$ & .82 $\uparrow^{\ast}$
& .79 & .65 $\uparrow^{\ast}$ & .72 $\uparrow$ \\
\cmidrule(l){2-14}
& LEXICON
& .90 & .88  & .89 $\uparrow$ 
& .89 & .85 $\uparrow^{\ast}$ & .87 $\uparrow$ 
& .79 $\uparrow^{\ast}$ & .87 $\uparrow^{\ast}$ & .85 $\uparrow^{\ast}$
& .79 & .77 $\uparrow^{\ast}$ & .81 $\uparrow^{\ast}$ \\
\specialrule{0.1em}{0.5em}{0.5em}
\multirow{4}{*}{GPT}
& BASE
& .80 & .53 & .67
& .73 & .31 & .52
& .36 & .19 & .28
& .47 & .24 & .36 \\
\cmidrule(l){2-14}
& IHRA
& .73 & .40 & .57
& .67 & .23 $\downarrow$ & .45
& .37 & .19 & .28
& .45 & .15 $\downarrow$ & .30 \\
\cmidrule(l){2-14}
& STRUCT
& .81 & .66 & .74
& .75 & .42 $\uparrow$ & .59 $\uparrow$
& .56 $\uparrow$ & .53 $\uparrow^{\ast}$ & .55 $\uparrow^{\ast}$
& .53 & .39 $\uparrow$ & .46 $\uparrow$ \\
\cmidrule(l){2-14}
& STRUCT+EX
& .78 & .74 & .76
& .72 & .47 $\uparrow$ & .60 
& .60 $\uparrow$ & .61 $\uparrow^{\ast}$ & .61 $\uparrow^{\ast}$
& .52 & .46 $\uparrow$ & .49\\
\specialrule{0.1em}{0.5em}{0.5em}
\multirow{4}{*}{LLaMA}
& BASE
& .74 & .71 & .73
& .62 & .32 & .47
& .38 & .27 & .33
& .40 & .18 & .29 \\
\cmidrule(l){2-14}
& IHRA
& .86 $\uparrow$ & .79 & .83 $\uparrow$
& .80 $\uparrow$ & .58 $\uparrow^{\ast}$ & .69 $\uparrow$
& .59 $\uparrow$ & .57 $\uparrow^{\ast}$ & .58 $\uparrow^{\ast}$
& .64 $\uparrow$ & .36 $\uparrow$ & .50 $\uparrow$ \\
\cmidrule(l){2-14}
& STRUCT
& .87 $\uparrow$ & .87 & .87 $\uparrow$
& .80 $\uparrow$ & .68 $\uparrow^{\ast}$ & .74 $\uparrow^{\ast}$
& .69 $\uparrow^{\ast}$ & .69 $\uparrow^{\ast}$ & .69 $\uparrow^{\ast}$
& .66 $\uparrow^{\ast}$ & .46 $\uparrow^{\ast}$ & .56 $\uparrow^{\ast}$ \\
\cmidrule(l){2-14}
& STRUCT+EX
& .87 $\uparrow$ & .84 & .86 $\uparrow$
& .77 $\uparrow$ & .61 $\uparrow^{\ast}$ & .69 $\uparrow$
& .62 $\uparrow$ & .62 $\uparrow^{\ast}$ & .62 $\uparrow^{\ast}$
& .62 $\uparrow$ & .40 $\uparrow$ & .51 $\uparrow$ \\
\bottomrule
\end{tabular}
\end{table*}

\begin{table*}[t]
\centering
\small
\footnotesize
\setlength{\tabcolsep}{2pt}
\caption{Comparison of annotated and referenced concepts. \textbf{Number of References}: Average number of \textit{Lexicon}-chapters/\textit{IHRA}-sections and content group referenced by model (row 1) or annotated by humans (row 2, in parentheses) per post. \textbf{Performance}: 
precision/recall/F$_1$ (row-wise) based on comparing content groups referenced in model explanations against human-annotated content groups, overall and per content group, and dataset (B = Bloomington, D = Decoding). Annotators assigned on average 1.3 section references (1.2 content groups references) per post for B and 1.48 chapter references (1.22 content group references) for D.}
\label{tab:rq3_distribution}
\begin{tabular}{l l cc!{\hspace{9pt}}cc || ll  !{\hspace{9pt}} ll !{\hspace{9pt}} ll !{\hspace{9pt}} ll !{\hspace{9pt}} ll}
\toprule
\multirow{3}{*}{\textbf{Model}} & \multirow{3}{*}{\textbf{Cfg}}
& \multicolumn{4}{c||}{\makecell{\textbf{Number of}\\\textbf{References}}}
& \multicolumn{10}{c}{\textbf{Performance}} \\
\cmidrule(lr){3-6} \cmidrule(lr){7-16}
& & \multicolumn{2}{l}{\makecell{\textbf{Sections/}\\\textbf{Chapters}}} & \multicolumn{2}{l||}{\makecell{\textbf{Content}\\\textbf{groups}}} & \multicolumn{2}{l}{\textbf{Overall}} & \multicolumn{2}{l}{\makecell{\textbf{Aggressive}\\\textbf{+Hate}}} & \multicolumn{2}{l}{\makecell{\textbf{Classic}\\\textbf{+Power}}} & \multicolumn{2}{l}{\makecell{\textbf{Israel}\\\textbf{-related}}} & \multicolumn{2}{l}{\makecell{\textbf{Post}\\\textbf{-Holocaust}}} \\
\cmidrule(lr){3-4} \cmidrule(lr){5-6} \cmidrule(lr){7-8} \cmidrule(lr){9-10} \cmidrule(lr){11-12} \cmidrule(lr){13-14} \cmidrule(lr){15-16}
& & B & D & B & D & B & D & B & D & B & D & B & D & B & D \\
\specialrule{0.08em}{0.3em}{0.3em}

\multirow{9}{*}{Claude}
& \multirow{3}{*}{IHRA}
& \multirow{3}{*}{\shortstack{2.00\\\textit{(1.32)}}} & \multirow{3}{*}{\shortstack{2.12\\\textit{(1.60)}}}
& \multirow{3}{*}{\shortstack{1.59\\\textit{(1.21)}}} & \multirow{3}{*}{\shortstack{1.72\\\textit{(1.27)}}}
& .70 & .66 & .38 & .19 & .52 & .65 & .89 & .77 & .36 & .32 \\
& & & & & & .86 & .84 & .44 & .74 & .94 & .84 & .93 & .93 & .62 & .39 \\
& & & & & & .81 & .79 & .41 & .30 & .67 & .73 & .91 & .84 & .46 & .35 \\
\cmidrule(l){2-16}

& \multirow{3}{*}{STRUCT}
& \multirow{3}{*}{\shortstack{3.30\\\textit{(1.33)}}} & \multirow{3}{*}{\shortstack{3.50\\\textit{(1.58)}}}
& \multirow{3}{*}{\shortstack{1.89\\\textit{(1.22)}}} & \multirow{3}{*}{\shortstack{1.86\\\textit{(1.27)}}}
& .63 & .66 & .35 & .18 & .52 & .67 & .91 & .76 & .16 & .35 \\
& & & & & & .90 & .87 & .76 & .87 & .91 & .72 & .93 & .97 & .80 & .78 \\
& & & & & & .76 & .77 & .48 & .30 & .66 & .69 & .92 & .85 & .27 & .48 \\
\cmidrule(l){2-16}

& \multirow{3}{*}{STRUCT+EX}
& \multirow{3}{*}{\shortstack{3.03\\\textit{(1.33)}}} & \multirow{3}{*}{\shortstack{3.22\\\textit{(1.60)}}}
& \multirow{3}{*}{\shortstack{1.75\\\textit{(1.23)}}} & \multirow{3}{*}{\shortstack{1.72\\\textit{(1.27)}}}
& .66 & .73 & .38 & .27 & .50 & .65 & .93 & .82 & .22 & .55 \\
& & & & & & .88 & .92 & .64 & .84 & .93 & .90 & .91 & .95 & .70 & .74 \\
& & & & & & .78 & .83 & .48 & .41 & .65 & .75 & .92 & .88 & .33 & .63 \\

\specialrule{0.08em}{0.3em}{0.3em}

\multirow{12}{*}{Gemini}
& \multirow{3}{*}{IHRA}
& \multirow{3}{*}{\shortstack{1.65\\\textit{(1.32)}}} & \multirow{3}{*}{\shortstack{1.62\\\textit{(1.60)}}}
& \multirow{3}{*}{\shortstack{1.51\\\textit{(1.22)}}} & \multirow{3}{*}{\shortstack{1.52\\\textit{(1.26)}}}
& .72 & .64 & .29 & .12 & .62 & .71 & .93 & .82 & .38 & .24 \\
& & & & & & .83 & .72 & .53 & .67 & .86 & .70 & .87 & .79 & .52 & .27 \\
& & & & & & .82 & .80 & .37 & .20 & .72 & .70 & .90 & .80 & .44 & .25 \\
\cmidrule(l){2-16}

& \multirow{3}{*}{STRUCT}
& \multirow{3}{*}{\shortstack{2.52\\\textit{(1.32)}}} & \multirow{3}{*}{\shortstack{2.60\\\textit{(1.51)}}}
& \multirow{3}{*}{\shortstack{1.58\\\textit{(1.21)}}} & \multirow{3}{*}{\shortstack{1.50\\\textit{(1.24)}}}
& .72 & .69 & .34 & .15 & .60 & .73 & .95 & .74 & .21 & .44 \\
& & & & & & .85 & .77 & .64 & .68 & .80 & .52 & .93 & .96 & .78 & .66 \\
& & & & & & .82 & .83 & .44 & .25 & .69 & .61 & .94 & .84 & .33 & .53 \\
\cmidrule(l){2-16}

& \multirow{3}{*}{STRUCT+EX}
& \multirow{3}{*}{\shortstack{2.10\\\textit{(1.32)}}} & \multirow{3}{*}{\shortstack{2.20\\\textit{(1.54)}}}
& \multirow{3}{*}{\shortstack{1.38\\\textit{(1.22)}}} & \multirow{3}{*}{\shortstack{1.42\\\textit{(1.25)}}}
& .73 & .76 & .47 & .23 & .56 & .74 & .95 & .80 & .29 & .59 \\
& & & & & & .80 & .83 & .49 & .63 & .81 & .74 & .86 & .92 & .69 & .63 \\
& & & & & & .85 & .86 & .48 & .34 & .66 & .74 & .90 & .86 & .41 & .61 \\
\cmidrule(l){2-16}

& \multirow{3}{*}{LEXICON}
& \multirow{3}{*}{\shortstack{2.35\\\textit{(1.32)}}} & \multirow{3}{*}{\shortstack{2.62\\\textit{(1.53)}}}
& \multirow{3}{*}{\shortstack{1.53\\\textit{(1.22)}}} & \multirow{3}{*}{\shortstack{1.62\\\textit{(1.25)}}}
& .67 & .73 & .47 & .25 & .48 & .67 & .93 & .79 & .22 & .57 \\
& & & & & & .80 & .89 & .44 & .84 & .85 & .85 & .87 & .93 & .63 & .74 \\
& & & & & & .82 & .83 & .45 & .39 & .61 & .75 & .90 & .85 & .33 & .64 \\

\specialrule{0.08em}{0.3em}{0.3em}

\multirow{9}{*}{GPT}
& \multirow{3}{*}{IHRA}
& \multirow{3}{*}{\shortstack{1.41\\\textit{(1.33)}}} & \multirow{3}{*}{\shortstack{1.43\\\textit{(1.66)}}}
& \multirow{3}{*}{\shortstack{1.25\\\textit{(1.24)}}} & \multirow{3}{*}{\shortstack{1.30\\\textit{(1.29)}}}
& .83 & .75 & .64 & .35 & .70 & .77 & .94 & .83 & .44 & .35 \\
& & & & & & .84 & .75 & .50 & .64 & .82 & .75 & .90 & .79 & .44 & .34 \\
& & & & & & .87 & .86 & .56 & .45 & .76 & .76 & .92 & .81 & .44 & .34 \\
\cmidrule(l){2-16}

& \multirow{3}{*}{STRUCT}
& \multirow{3}{*}{\shortstack{3.14\\\textit{(1.35)}}} & \multirow{3}{*}{\shortstack{3.07\\\textit{(1.58)}}}
& \multirow{3}{*}{\shortstack{1.92\\\textit{(1.24)}}} & \multirow{3}{*}{\shortstack{1.94\\\textit{(1.27)}}}
& .61 & .60 & .29 & .12 & .51 & .51 & .96 & .83 & .21 & .45 \\
& & & & & & .88 & .85 & .86 & .86 & .85 & .70 & .91 & .92 & .75 & .73 \\
& & & & & & .74 & .73 & .43 & .21 & .64 & .59 & .93 & .87 & .33 & .56 \\
\cmidrule(l){2-16}

& \multirow{3}{*}{STRUCT+EX}
& \multirow{3}{*}{\shortstack{2.81\\\textit{(1.34)}}} & \multirow{3}{*}{\shortstack{2.73\\\textit{(1.61)}}}
& \multirow{3}{*}{\shortstack{1.86\\\textit{(1.23)}}} & \multirow{3}{*}{\shortstack{1.76\\\textit{(1.28)}}}
& .61 & .67 & .35 & .22 & .43 & .50 & .97 & .88 & .23 & .63 \\
& & & & & & .87 & .89 & .76 & .79 & .91 & .85 & .89 & .91 & .64 & .72 \\
& & & & & & .75 & .78 & .48 & .34 & .58 & .63 & .93 & .89 & .34 & .67 \\

\specialrule{0.08em}{0.3em}{0.3em}

\multirow{9}{*}{LLaMA}
& \multirow{3}{*}{IHRA}
& \multirow{3}{*}{\shortstack{1.75\\\textit{(1.32)}}} & \multirow{3}{*}{\shortstack{2.07\\\textit{(1.52)}}}
& \multirow{3}{*}{\shortstack{1.50\\\textit{(1.21)}}} & \multirow{3}{*}{\shortstack{1.78\\\textit{(1.24)}}}
& .70 & .57 & .51 & .25 & .55 & .62 & .92 & .74 & .18 & .14 \\
& & & & & & .81 & .76 & .35 & .51 & .84 & .71 & .89 & .85 & .60 & .46 \\
& & & & & & .82 & .75 & .42 & .34 & .66 & .66 & .90 & .79 & .28 & .21 \\
\cmidrule(l){2-16}

& \multirow{3}{*}{STRUCT}
& \multirow{3}{*}{\shortstack{2.99\\\textit{(1.34)}}} & \multirow{3}{*}{\shortstack{3.24\\\textit{(1.50)}}}
& \multirow{3}{*}{\shortstack{1.74\\\textit{(1.23)}}} & \multirow{3}{*}{\shortstack{1.76\\\textit{(1.24)}}}
& .64 & .63 & .33 & .17 & .51 & .64 & .92 & .70 & .14 & .31 \\
& & & & & & .83 & .82 & .35 & .64 & .89 & .68 & .91 & .95 & .74 & .64 \\
& & & & & & .77 & .77 & .34 & .27 & .65 & .66 & .91 & .81 & .24 & .42 \\
\cmidrule(l){2-16}

& \multirow{3}{*}{STRUCT+EX}
& \multirow{3}{*}{\shortstack{2.49\\\textit{(1.36)}}} & \multirow{3}{*}{\shortstack{2.62\\\textit{(1.52)}}}
& \multirow{3}{*}{\shortstack{1.56\\\textit{(1.24)}}} & \multirow{3}{*}{\shortstack{1.60\\\textit{(1.24)}}}
& .68 & .68 & .39 & .20 & .54 & .68 & .93 & .73 & .17 & .42 \\
& & & & & & .81 & .82 & .45 & .68 & .85 & .68 & .86 & .94 & .58 & .61 \\
& & & & & & .80 & .81 & .42 & .31 & .66 & .68 & .89 & .82 & .26 & .50 \\

\bottomrule
\end{tabular}
\end{table*}
\end{document}